# Dutch Comfort:

# The limits of AI governance through municipal registers

**Authors:** Corinne Cath[1,2,] and Fieke Jansen[3]

[1] Oxford Internet Institute, University of Oxford, 1 St Giles, Oxford, OX1 3JS, United Kingdom.
[2] The Alan Turing Institute, headquartered at the British Library, 96 Euston Road, London NW1 2DB, United Kingdom. Author email: ccath@turing.ac.uk
[3] Data Justice Lab, School of Journalism, Media and Culture, Cardiff University, Two Central Square, Central Square, Cardiff CF10 1FS, United Kingdom. Author email: jansenf@cardiff.ac.uk

**Acknowledgments:** We would like to thank Os Keyes, Janet Haven, and Michael Veale for their feedback on early drafts of this research. We are grateful for our ongoing discussions with the Amsterdam Municipality and Dutch civil society, and their willingness to discuss current attempts in The Netherlands to provide accountability for AI systems. Last but not least, we are grateful to our research funder: the ERC Starting Grant 'Data Justice: Understanding datafication in relation to social justice' (DATAJUSTICE) (no. 759903).

# Abstract

In this commentary, we respond to a recent editorial letter by Professor Luciano Floridi entitled "AI as a public service: Learning from Amsterdam and Helsinki". Here, Floridi considers the positive impact of these municipal AI registers, which collect a limited number of algorithmic systems used by the city of Amsterdam and Helsinki. There are several assumptions about AI registers as a governance model for automated systems that we seek to question. Starting with recent attempts to normalize AI by decontextualizing and depoliticizing it, which is a fraught political project that encourages what we call 'ethics theater' given the proven dangers of using these systems in the context of the digital welfare state. We agree with Floridi that much can be learned from these registers about the role of AI systems in municipal city management. Yet, the lessons we draw, on the basis of our extensive ethnographic engagement with digital well-fare states are distinctly less optimistic.

# 1. Introduction

Many recent academic publications about the use of technology in modern society—from the critical, like urban infrastructure, law enforcement, banking, healthcare, and humanitarian aid, to the mundane, like dating—have taken up the siren call of Artificial Intelligence (AI). AI is a catch-all phrase for a wide-ranging set of technologies, most of which apply learning techniques from statistics to find patterns in large sets of data and make predictions based on those patterns. Various academics argue that AI holds great promise for the economy, social welfare, and urban development, but it can also act unpredictably and do great harm (Cath, 2018; Rincón, Keyes, & Cath, 2021; Veale & Binns, 2017). We argue that these harms often arise from a lack of contextual awareness and a tech-deterministic view of the role AI could, and should, play in the provision of services, including by the state. Our findings are based on our respective Ph.D. research projects and fieldwork in The Netherlands.

In this commentary, we focus on the recent emergence of municipal "AI registers" in the context of Amsterdam and Helsinki, with a distinct focus on the Dutch case. These municipal AI registers are websites that aim to offer a repository of algorithms deployed within a specific geographical area, including a limited description of how and why they are used. Currently, the Helsinki register[1] holds five entries that primarily concern the use of chatbots within the context of municipal service provision. The Amsterdam register[2] holds three entries that span a broad range of algorithmic function, including the use of an automated parking control algorithm through object detection, keyword detection on complaints about issues in public space which allows the municipality to funnel the report more effectively to the right department, and an algorithm that calculates the probability of individuals illegally renting out their properties for holiday purposes. Or in other words, the entries in these registers are narrowly scoped to cover a set of largely uncontentious bureaucratic municipal uses of AI.

Principally, in this commentary we argue that AI registers must be understood in their embedded context and as tools that consolidate existing information, and thus power, inequities between governments and city dwellers. Our argument aligns with Joseph Weizenbaum (1985) notion of computation as a conservative force, which is used to describe how the computer has allowed institutions to keep and even solidify power. Computers enable governments to optimize specific bureaucratic challenges that might have otherwise required them to more fundamentally change how they operate. There is a direct parallel between this notion of the computer as a conservative force and how municipal AI registers, and the accompanying discourse surrounding ethical safeguards, are used to normalize the fraught political project of urban AI.

---

[1] https://ai.hel.fi/en/ai-register/
[2] https://algoritmeregister.amsterdam.nl/en/ai-register/

We make this specific argument in response to a recent editorial letter by Professor Luciano Floridi (2020) about "AI as a public service: Learning from Amsterdam and Helsinki". In his letter, published less than a month after these two municipalities announced the launch of their "AI registers", Floridi praises these databases as a vehicle to normalize AI. He argues that "the value of the project seems to lie in the recognition that AI as a "utility" is a great means to deal with increasingly complex, urban environments" (2020, p. 542). Throughout his letter, Floridi highlights the positive features of the registers, in a way that uncritically assumes AI should be implemented as a public service in cities. His "gaze from nowhere" (Haraway, 1990) inscribes specific judgments and values to both AI and its emerging governance processes through which he neutralizes social justice and governance concerns to that of the merely technical (Crawford, 2018).

Rather than taking an idealized philosophy of information approach to these AI registers, our approach is rooted in Science and Technology Studies (STS) as well as critical social sciences. In responding to Floridi's letter, we articulate two fundamental concerns arising out of his appraisal of the registers, namely the dangers of romanticizing the register as a governance solution and normalizing the use of AI in cities. We demonstrate that a lack of contextualization can lead to hastily lauding these voluntary databases for functions that do not address the most pressing problems arising out of urban use of AI systems. Superficial consideration of AI registers' efficacy leads academics to conflate their existence with the utility of AI systems for urban management. At the same time, surface-level engagement with these proposed solutions can inflate the utility of voluntary localized databases for the governance of technology. These two concerns lead us to the articulation of a transcendent concern in our discussion section, in which we question the assumption that cities need AI.

We argue that, while a laudable first step towards accountable and locally driven AI governance, we must not romanticize registers. Doing so overlooks how much information is currently *not* included in these databases. For both STS academics and infosphere theorists, the absence (or destruction) of information is seen as a decisive ethical act. Evidently missing from the Amsterdam and Helsinki examples, are some of the most harmful applications of AI in welfare and law enforcement. While we agree with Floridi that "additional [AI] services can easily be added" (2020, p. 544), we outline several political realities that make it unlikely that the most sensitive of applications will be. These political realities informing the informational voids in the register are as telling, if not more than the data they contain.

Articulating this standpoint is critical against the broader backdrop of current discussions about AI governance and regulation in Europe. In late April 2021, the European Commission published its long-awaited legal framework for regulating AI in the European Union (EU). This text provides the basis for European negotiations about the final shape of this regulation, which will take place over the next few years. While the EU's AI framework contains multiple stipulations regarding the registration of AI systems on the EU market in open databases (Veale & Zuiderveen Borgesius, 2021), not unlike the municipal registers at the center of this paper, the Commission's approach is primarily one that aims to create an enabling environment for a European AI market to develop (Jansen, 2021). This context helps us position these municipal AI registers, and Floridi's editorial letter, in a wider societal debate. This debate is often overly focused on creating the conditions that allow AI to be deployed across society, rather than critically engaging with the question of whether AI systems should be deployed in the first place. Our critique also provides context to the limited ability of ex-ante databases to safeguard rights and justice.

In The Netherlands in particular, as we will outline throughout this piece, there has been much public commotion about discriminatory automated welfare distribution systems (Roosen, 2020) and police use of AI (Amnesty International, 2020). In 2021, the Dutch government fell over a child benefit scandal where the tax office relied on automated systems that wrongly accused thousands of families of well-fare fraud. Here, using AI systems, allows public authorities, from the tax office to the police, to sanitize their organizational choices from the politics of who is perceived, or constructed, as primary agents of welfare fraud or crime. These often racialized and class constructs determine which communities are subjected to increased government scrutiny and oppression (Williams & Clarke, 2018; Browne, 2015), and hardwire the logic of the repressive welfare states in technology (Vonk, 2014). These controversies show how registers are clearly only a partial solution to the many governance concerns arising from the growing use of AI systems, not least those related to public accountability.

This commentary is organized as follows. First, we provide a summary of the Dutch AI register program and Floridi's light touch consideration of it. Subsequently, we will analyze the AI registers within the context of the politics of digital technology in The Netherlands. After which, we will outline two concerns with Floridi's detached description of the AI registers: the dangers of romanticizing the register and normalizing AI. We use these dual concerns to discuss the overarching question left unspoken, and unanswered, by the proponents of AI registers: should governments, in this case municipalities, be deploying AI systems at all? We conclude by outlining the twin dangers raised by non-empirical analysis of AI registers: it legitimizes the use of AI systems in the provision of public services and fails to acknowledge which ideologies, inequities, and political rationales are embedded within a municipal turn to AI. In doing so, we argue that registers reflect the dangers of embedding AI systems in the fabric of city life, as much as they present a novel and accountable approach to AI governance.

## 2. The Unbearable Lightness of AI Registers in Amsterdam and Helsinki

In September 2020, the cities of Helsinki and Amsterdam provided a novel answer to ongoing discussions about how to ensure accountability for governmental use of AI systems in cities. These two European hubs did so by setting up open registers, publicly accessible databases that provide an overview of AI systems and algorithms used by the cities for a variety of purposes. The AI registers, in their current incarnation, are unbearably light. That is to say, they provide a limited overview of the algorithmic systems used and developed by the cities on an opt-in basis. At the time of writing, the Amsterdam and Helsinki register contain only eight entries in total.[3] The entries mostly cover automated government services, including city libraries and hospitals. The registers do not contain any corporate sector entries or information about algorithmic systems used in critical governance areas like law enforcement or welfare provision. Yet, it is these sectors that are often implicated in algorithmic discrimination (Benjamin, 2019; Eubanks, 2019). This exclusion means that some of the most sensitive applications of AI are not currently covered by the registers, nor are they likely to be included in the future.

---

[3] Interestingly, the Dutch register over the course of the process of writing and getting this commentary published between January and August 2021, removed two entries in the register. Including one that was focused on automated GAIT recognition to enforce the 1,5-meter distance measure during peaks in the COVID-19 pandemic. The entry can still be found on the website, but not on the homepage. It is unclear what prompted the removal of these entries, further stressing our point that these registers will struggle to provide accountability of municipal use of AI systems.

We raise a number of critical questions about registers as governance solutions by responding to various concerns arising from Floridi's commentary on the importance of these novel AI registers. In his 2020 editor's letter published just after the announcement of the AI registers, Floridi describes the registers by discussing their positive impact. In particular, he argues that negative connotations about AI are preventing society from harnessing this technology's potential. Or in Floridi's words, we should view AI as "a new form of mindless agency into which one can tap to deal with problems that otherwise would require human intelligence and perhaps a huge (sometimes unfeasible) amount of other resources" (2020, p. 542). Given the limited and voluntary remit of the current registers, Floridi's analysis of their role in "normalizing AI as just another public utility" and providing public accountability by "creating a culture in which it is the human users who can watch and hence determine the behaviour of AI services" is premature. This analysis omits acknowledgment of the limitations and failures of similar initiatives in cities like New York (Lecher, 2019). The promise of bottom-up oversight, however, helps governments and other deployers to ensure the technology becomes viewed as a mere utility that allows for more efficient and effective service provision in urban areas. This same line of thinking is what enables regulators to prop up AI registers as vehicles for achieving transparency and thereby trust. Rather than using the register to challenge the use of AI, or question those in positions of power who position AI deployment as inevitable, it becomes a vehicle to normalize it by demonstrating its benign utility.

Instead of reflecting on the potential positive impact of the register, which we consider to be too early to assess, we situate these registers in the context of local power dynamics and through the lens of critical AI scholarship. This allows us to question several of Floridi's assumptions, especially attempts to decontextualize, depoliticize, and normalize AI, which in itself is a questionable academic pursuit given recent adverse uses of AI systems by various governments (Algorithm Watch, 2020; Dencik et al, 2018), including the Dutch government (Privacy International, 2020). We agree with Floridi, and other commentators, that much can be learned from these registers about the role of AI systems in municipal city management. Yet, the lessons we draw from the Amsterdam register, on the basis of our extensive ethnographic engagement with digital well-fare states and our first-hand experience with the Dutch context, are distinctly less optimistic. In order to do so, we will first contextualize the use of AI systems in The Netherlands with a focus on its most famous city.

## 3. The Politics of Digital Technology in The Netherlands

The Netherlands was an early Internet adopter, its first connection to the Internet's predecessor was established in 1988 (Franx, 2018). As early as 1993, commercial Internet Service Providers (ISP) started widely accessible connections. The long history of digital technology adoption in The Netherlands developed in unison with civil society critique of its potentially negative impacts on fundamental rights and liberties. Much of the early online community grew out of hacking and anarchist collectives throughout the country. Only a decade after John Perry Barlow's famous "Declaration of the Independence of Cyber Space" (1996), a small group of digital rights activists founded "Bits of Freedom". This Dutch digital rights organization subsequently helped found the European Digital Rights Initiative (EDRi): one of the leading voices on issues of privacy, surveillance, and tech harms in Europe.

The long history of early technology adoption, as well as critical civil society interventions, by the Dutch extends into the era of AI. Most recently a collective of activists and strategic litigators[4] made world news when they won a court case against the Dutch government's use of an automated system for welfare fraud detection. The collective successfully argued that this surveillance system was stacked against the poor and presented a violation of privacy laws. The judge's verdict was damning. It explicitly mentioned that the use of the system was not only in disagreement with local laws but also with international human rights standards, making this verdict relevant to global efforts at digitizing welfare states.

A crucial component of this case was the black-box nature of the fraud detection system. The Dutch government provided little information about its inner workings, raising obvious accountability concerns. While this system predates Amsterdam's register, given the government's secrecy it is unlikely that this fraud detection system or others like it, would have been included in the register. Even if it would have been included, this would have still left many of the other cities in which this detection system was trialed in the dark. The Amsterdam AI register is thus not a comprehensive answer to ensuring public accountability for the application of algorithmic systems. The registers do not reckon with the various cultural, political, and economic incentives that steer the use of automated systems for governmental services and goods provision.

---

[4] This collective includes the Public Interest Litigation Project (PILP),
Stichting Platform Bescherming Burgerrechten, Dutch Section of the International Commission of Jurists (NJCM), Union FNV, Stichting Privacy First, Stichting KDVP and the Landelijke Cliëntenraad) and authors Tommy Wieringa and Maxim Februari. More about this collective can be found here:
https://pilpnjcm.nl/en/proceedings-risk-profiling-dutch-citizens-syri/

Drawing back on the work of Floridi to which we are responding, we note the lack of an in-depth exploration of the Dutch state's impetus to automate its services. Such an inquiry is necessary to show that the use of AI systems does not reflect an inevitable step in urban development based on the idea that "with enough data, statistics, and computational power, one can do without intelligence" (Floridi, 2020, p. 541). Rather the use of these systems is calculated and reflects a myriad of social pressures facing the Dutch government. Two such pressures came up as crucial in our research of Dutch digitization efforts. First, like in many other social welfare states in Europe, The Netherlands has seen a succession of government coalitions led by right-wing parties. Their agendas are often based on a neoliberal logic, which posits that the size and role of the state should be reduced, and welfare services should only be accessible to those who qualify as 'deserving' poor (Wacquant, 2009). This logic is coupled with a belief that outsourcing and automating services increase government efficiency. The managerial logic of effectiveness and efficiency, as multiple scholars looking at the impact of AI systems find, tends to have a disproportionate impact on the most marginalized in society (Eubanks, 2019). Second and specific to the Dutch context, we found that a "just do it" culture guides government organizations and institutes tasked with upholding the law, including the police.

A contextual understanding of technological deployment allows us to position the government and city interest in AI within their political ideologies and socio-technical imaginaries. In the Dutch case, the political belief in AI's potential as a neutral utility is connected to the perception that public service provision is devoid of class and race structures. Approaching these registers in connection with their broader political context allows us to see how they are shaped by, and in return shape, the perceived legitimacy of AI and the governance mechanisms that surround them. Throughout the remainder of this commentary, we stress the importance of these political pressures in understanding the limited utility of AI registers, in their current incarnation. In the next sections, we respond to two fundamental concerns arising out of Floridi's analysis, by providing a rebuttal to his underlying assumptions about the potency of AI. We also highlight the hazards of romanticizing the register as a way to minimize the harmful impact of AI.

## 4. Premature Romanticizing of AI Registers

In reflecting on Floridi's editorial letter, we must note that despite his valid critique of debates about AI being overly focused on "pointless but very distracting speculations about nasty robots, singularity, superintelligence, and other sci-fi dystopian stories" (2020, p. 541), his arguments move to the opposite extreme. Instead of demonizing AI, he romanticizes the light touch ex-post transparency measures provided by the AI registers in Helsinki and Amsterdam. We counter this romanticized notion of the registers by arguing that any analysis of its merits should not solely look at what it aspires to be but what it is, and more importantly at what it is not. We will probe Floridi's excitement about the emergence of AI registers by exploring its current information voids.

For those familiar with the context of Amsterdam, and the extent to which algorithms are already used in Dutch 'public space' and by public authorities, it is clear that the algorithmic register only reveals those systems that are housed by the municipality. It lacks information about systems deployed within the city limits by other public authorities or third parties. The automated parking control algorithm is one of the three entries[5] in the Amsterdam database. Yet, this entry provides only a partial picture of the use of mobile sensing vehicles operational within the city, as it excludes the various commercially operated scanning vehicles. The object detection entry[6], in which the city is experimenting with GAIT recognition for crowd monitoring purposes does not account for the police facial recognition trials in these same locations (Amsterdam Smart City Chief Technology Office, 2019). The register might aim to decrease the information asymmetry between citizens and the municipality, but it will likely increase this asymmetry if it precludes corporate or contentious government systems.

---

[5] This entry and the other examples can be found on the city's "Algorithm Register" website here: https://algoritmeregister.amsterdam.nl/en/automated-parking-control/

[6] See the one-and-a-half-meter monitor entry here: https://algoritmeregister.amsterdam.nl/en/one-and-a-half-meter-monitor/ Interestingly, somewhere over the summer of 2021 the municipality removed this entry from the homepage of the website. See also footnote 3.

As it stands, this transparency effort might obfuscate AI systems used by other public authorities, like the police and judiciary. It also conceals that the City of Amsterdam is not only an operator of AI for public services. The city, at times, shares and buys data from commercial parties that deploy algorithms in public spaces (Munten, 2019). Here, we refer to the trend whereby public authorities become increasingly dependent on, while also voluntarily outsourcing, the collection and processing of data by commercial entities. This growing trend further entrenches "a dependency on an economic model that perpetuates the circulation of data accumulation" (Dencik, 2021). Engaging with what is, rather than then what is not, overlooks that these registers offer a technocratic snapshot in time of a small subsection of algorithms deployed by or available to public authorities. This overestimates the role of AI registers, which are but nascent transparency initiatives.

These debates about the contours of accountability protections provided through open data are not new; social scientists have long romanticized and problematized the relationship between transparency and accountability (Gaventa & McGee, 2013; Grimmelikhuijsen, 2012; Lourenço et al, 2017). Where it is generally understood that the first does not imply the latter, transparency is a starting point, a tool that might in the best of circumstances lead to more accountability. Yet it often happens at the hands of a small group of underfunded and overburdened experts, civil society organizations and investigative journalists. These insights notwithstanding, Floridi ends his editorial letter with the argument that, "The fact that someone may be watching is sufficient to make a positive difference in terms of what should and should not be done, to begin with, when it comes to using AI as a public service" (2020, p. 545). Here, the promise of the register to inform the citizens provides the basis for Floridi's lessons, as such rejecting ample scientific evidence that "watching the watcher" (Monahan, 2006; Welch, 2011) is often insufficient to hold those in power to account. It also discounts the ongoing efforts of critical voices in Dutch society that predate the register.

The individuals who were already watching, investigating, challenging, and questioning the use of algorithms in public spaces and public services before the register was even conceptualized, the authors of this piece included, are skeptical that it will actually offer sufficient information needed for robust accountability efforts. From our conversations with individuals working for the Amsterdam municipality, it became clear they are aware of these limitations. Their honesty makes academic ruminations, especially those painting an optimistic image of the register that attributes possibilities and mandates for the sake of argument without acknowledging these known limitations, moot. Academic inflations of the register can contribute to harmful false narratives about the use, utility, and risks of algorithms for the 'public good' and suitable governance mechanisms, what we call 'ethics theater'. The act of merely focussing on the potential of visible, but limited, transparency measures and not engaging with its context nor underlying power relations can be compared with the phenomenon of security theater. In security theater, visible security measures aim to provide a sense of security, but in reality, provide little or no actual security (Schneier, 2003; Johnston & Warner, 2010). In the next section, we elaborate on these findings by questioning who benefits from romanticizing AI (and their respective registers) against the backdrop of bureaucratic violence in The Netherlands.

## 5. The Dangers of Normalizing AI

The register acts as a political project to normalize the use of AI systems for society critical government services, like urban management and municipal affairs. As mentioned, both academic and practitioner proponents of the register are active participants in this project. At a fundamental level, proponents argue the register is meant to reduce societal fear of AI systems and consequently increase public trust in their use. While Floridi mentions the limits of trust through transparency alone (2020, p. 545) he does conceptualize the development of the Dutch register in terms of its ability to reduce societal anxieties around what he terms "fear that Terminator [sic] might be coming" (2020, p. 542). The Netherlands is certainly not immune to conspiracies about new technologies (RTL News, 2020) but the fear described by Floridi is a reductive take on legitimate Dutch concerns about AI systems. It posits the Dutch as simple technophobes, behind on the times who need their benevolent state to take them by the hand into the digital age, rather than acknowledging that these critiques center rights and well-being of individuals at the loci of struggle rather than trust in AI.

The previous section, however, indicated that the Dutch were early adopters of both emerging technologies and critical societal bottom-up responses to its uses by the government. Floridi's description of the Dutch context flattens the country's history. It also overlooks that the lawsuit against the Dutch government's use of automated welfare fraud detection was the first such comprehensive case worldwide to question the digital welfare state on human rights grounds (Alston, 2019). Describing Dutch fear in such "Schwarzeneggerian" terms diminishes justified societal concerns around government use of AI systems. This characterization is akin to describing resistance against the implementation of AI systems, rooted in ground truth about bureaucratic and state violence, as prompted by the fear of a fictional cyborg assassin like the Terminator. Yet, Floridi's description of AI as "normal, common, utility like" (2020, p. 543) overlooks that the mundane uses of simple robots or AI systems can also be violent, and it is not hyperbole to take issue with them.

Recent developments suggest that existing concerns around state use of AI in The Netherlands are justified: the earlier mentioned 2021 collapse of the Dutch government following the revelations of fundamental flaws in their adversarial use of an automated fraud detection system are but one example. United Nations' Special Rapporteur on extreme poverty Alston describes (2019, p. 7) the use of AI systems in The Netherlands as part of a political development in which technology is "mobilized to target disproportionately those groups that are already more vulnerable and less able to protect their social rights". This political context seems missing from Floridi's analysis, even though it is crucial to understand the power dynamics that shape how AI is seen and the efficacy of AI registers in The Netherlands. The fear for AI systems is thus squarely rooted in immediate concerns for human rights, rather than on an unfounded worry about robotic overlords.

The AI registers normalize the earlier mentioned neoliberal logic and "just do it" culture guiding Dutch politics. By defining accountability as voluntary registration, the state is essentially taking an ex-post approach to accountability. Use and develop the system first ("Just do it") and ask for forgiveness later. This approach has an eerie "Californian ideology" (Barbrook & Cameron, 1995) ring and reflects a fundamentally up-side-down approach to governance: as it conflates registering AI systems with enabling accountability for their material impact. It is dangerous to overlook the resistance against the normalization of AI as a value or power struggle that requires more political deliberation and instead see it as temporary frictions that can be mitigated through technocratic processes. This approach eschews difficult conversations about whether an AI system should be implemented *at all* or how it contributes to disenfranchising some while privileging others, as it assumes the damage has already been done. This ex-post AI governance approach also bypasses conversations about the root motivations for rolling out AI systems in government and who are truly served by them.

A cursory glance at recent work by prominent Dutch digital rights organizations reveals they are worried about the political impact of AI systems on equality, anti-discrimination, and privacy. For example: a coalition of digital rights organizations (including Bits of Freedom) launched an EU-wide campaign to "Reclaim Your Face"[7], which calls on a ban of facial recognition and biometric mass surveillance. These worries are justified, given the earlier mentioned example of the problematic welfare fraud detection system. Likewise, various Dutch political parties recently pushed back on government use of 'smart' algorithms. They fear it will disproportionately infringe on citizens' fundamental rights, arguing that the use of AI as a public service should be banned until there are clear rules and regulations that will govern its use (Schellevis, 2019). These organizations and individuals, hence, do not fear a fictitious Terminator, they fear the factual State and what it might do when AI is normalized. This brings us to the discussion of the fundamental concern arising from Floridi's examination of AI registers: the implicit assumption that cities should implement AI solutions at all.

## 6. Discussion: Do cities need AI?

The supposed importance of the role that AI could and should play in municipal level city management is stooled on the flawed assumption that both the technology and the city are neutral actors. The urban drive for AI registers is fuelled by a tech-deterministic philosophy that assumes the use of AI is inevitable. In the editorial letter, Floridi argues that "framing AI within the already available good practices of urban policies is refreshing" and "the [register] project shows daily realities and experiences in the life (or rather onlife) of ordinary citizens" (2020, p. 543). By ascribing neutrality to AI and focusing on registers as a best practice of governance, he isolates the city, its AI systems, and the register from questions of power.

From an international governance or information philosophy perspective, municipalities and AI might be just this, local benevolent actors. But from the perspective of its inhabitants, municipalities are state authorities who control different aspects of their lives. Data harms and inequality are not the mere result of technologies, they are the result of larger social power structures that cannot simply be optimized away by transparency through registers. Dismissing the idea that cities in themselves are a state authority that exerts power over its inhabitants and over other public authorities and commercial entities, allows proponents of AI to position the registers outside historic and ongoing power struggles over public spaces and services in Amsterdam.

---

[7] The full campaign and the various organizations involved in it can be found here: https://edri.org/our-work/campaign-reclaim-your-face-calls-for-a-ban-on-biometric-mass-surveillance/

This "artificial" separation of the city and AI from questions of power, in turn, discounts critical scholars that have exposed the discriminatory harms in the use of algorithms in public services (Collins, 1999; Williams & Clarke, 2016) as well as the experience of affected communities on the ground. Positioning any public authority or AI system as neutral, means disqualifying decades of social justice struggles across the world, in which anti-discrimination and critical race scholars (Benjamin, 2019; Eubanks, 2019; Keyes, 2018; Mohamed, Png & Isaac 2020; Raji, 2020; Richardson et al., 2019; Rincón, Keyes & Cath, 2020), civil society and activists have fought an uphill battle to create a vocabulary, awareness, and acknowledgement of institutional racism and its imbrication with technology. It is exactly this positioning of algorithms as a benevolent utility that allows Floridi to abstract generalized positive lessons from it. He makes democratic legitimacy and harms not about politics but about technocratic process, this creates a comfortable situation for him and other proponents of AI registers, in which they do not need to account for context, lived harms or for existing structures of oppression and domination (Dencik et al., 2018).

In his concluding remarks, Floridi argues that even if the registers are not successful, they "shows a very good way in which AI could be implemented as a public service, thus indicating a fruitful strategy that societies could and should pursue" (2020, p. 545). As mentioned before this is an ex-ante approach to governance, that conflates a register, a minimal and voluntary snapshot of the use of algorithmic systems by one public authority, with the actual systems that are deployed within public services. The register might aim to increase transparency about the uses of AI in public service, but it cannot be lauded for more than that. The actual algorithms predate the idea of register and have been shaped by the deeply rooted political ideology, organizational cultures, and interests that materialize in political decision-making. Conflating these two is the continuation of attempts to collapse the political into the technical, "as if the solution to societal and political conflicts were simply a matter of imperfect information" (Andrejevic, 2019, p. 101). As such this emphasis on transparency as a fruitful strategy that cities across the world should pursue, ex-post governance, obfuscates more fundamental questions about the purpose and necessity for AI as a public service in the first place.

## 7. Conclusion

We hope that the experiment in Amsterdam and Helsinki with the AI registers will be a success. For that to happen, further reflection on cities trialing AI governance-by-database must be rooted in the local social and political context. It also needs to consider the actual practices of those involved in their creation and acknowledge the ongoing critical engagement of civic actors on issues of AI governance, rather than mere 'ethics theater'. If that contextualization does not happen, it is unlikely that these AI registers will be able to offer more than the "Dutch comfort" that "things could be worse". We encourage wider debate on AI governance and automated public services that are rooted in the material and political possibilities and limitations of registers. Fomenting such debates requires us to understand the challenges practitioners face and how the use of algorithms transforms existing power structures within the state-market-citizen nexus.

In ongoing discussions about city futures, we must be willing to accept that AI systems are rarely the right means to deliver public services. The internal logic of efficiency and standardization of these systems is an awkward fit with the manifold and messy needs of humans, especially those at the margins of society. Automating government services means making the state more machinist and less approachable, or accountable. Defining AI in terms of public utilities implies there are no issues with the existing governance frameworks for other utilities like water and gas, a highly speculative assumption. Rather than being content with the examples set by Amsterdam and Helsinki, through this commentary, we want to encourage critical interrogation of the presumed relationship between AI registers, governance, and accountability. Databases are a part of the puzzle of AI governance, but one that only fits as part of a comprehensive political framework that clearly sets boundaries of use, and accountability standards, for both public and private use of automated systems.